\pdfoutput=1

\documentclass[11pt]{article}

\usepackage[]{acl}

\usepackage{times}
\usepackage{latexsym}
\usepackage{ragged2e}
\usepackage{graphicx}
\usepackage{caption}
\usepackage{xcolor}

\usepackage[T1]{fontenc}

\usepackage[utf8]{inputenc}

\usepackage{microtype}

\usepackage{inconsolata}

\title{Order-Based Pre-training Strategies for Procedural Text Understanding}

\author{Abhilash Nandy ~~~ Yash Kulkarni ~~~ Pawan Goyal ~~~ Niloy Ganguly \\ \texttt{nandyabhilash@kgpian.iitkgp.ac.in} \\ Indian Institute of Technology Kharagpur \\ India}

\begin{document}
\maketitle
\begin{abstract}

In this paper, we propose sequence-based pre-training methods to enhance procedural understanding in natural language processing. Procedural text, containing sequential instructions to accomplish a task, is difficult to understand due to the changing attributes of entities in the context. We focus on recipes, which are commonly represented as ordered instructions, and use this order as a supervision signal.
Our work is one of the first to compare several `order-as-supervision' transformer pre-training methods, including Permutation Classification, Embedding Regression, and Skip-Clip, and shows that these methods give improved results compared to the baselines and SoTA LLMs on two downstream Entity-Tracking datasets: NPN-Cooking dataset in recipe domain and ProPara dataset in open domain. Our proposed methods address the non-trivial Entity Tracking Task that requires prediction of entity states across procedure steps, which requires understanding the order of steps. These methods show an improvement over the best baseline by $1.6\%$ and $7$-$9\%$ on NPN-Cooking and ProPara Datasets respectively across metrics.\footnote{Code is available at \url{https://github.com/abhi1nandy2/Order_As_Supervision}}
\end{abstract}
\section{Introduction}
Procedural text comprises a series of sequential instructions aimed at guiding individuals through a task by presenting information in a step-by-step manner. A procedure describes a step-wise interaction between multiple participating entities and their attribute changes. For instance, "Photosynthesis" as a procedure consists of interaction between entities such as water, light, CO2, sugar, etc. Recently, there has been an increase in the number of studies in NLP that use procedural texts. Procedural text is common in natural language in recipes \cite{recipe1M_paper, bien-etal-2020-recipenlg, Chandu2019StoryboardingOR,bodhiswatta_paper, npn_paper}, how-to guides \cite{nandy-etal-2021-question-answering}, and scientific processes \cite{propara-paper}. In this study, we focus on recipes as they are commonly represented as ordered instructions. We utilize this order as a supervision signal to develop customized pre-training techniques to solve non-trivial tasks that require anticipating the implicit effects of actions on entities.

Understanding procedural text is difficult due to the changing attributes of entities in the context. Previous works such as \citet{slm} used Sentence-level Language Modeling (SLM) to learn contextualized sentence-level representation by training a hierarchical transformer to reconstruct the original order of a shuffled sequence, \citet{ien} proposed Interactive Entity Network (IEN) to model different types of entity interactions using a recurrent network with memory for state tracking, and \citet{Zhang_2021} combined external knowledge with a BERT model to improve entity tracking.  However, such works do not compare pre-training techniques which consider sequential order of the steps of the procedure. 


In this paper, we try to solve the non-trivial Entity Tracking Task that requires prediction of entity states across procedure steps. Solving such a task requires understanding the \textit{sequential nature/order of the steps}. Explicitly learning the order within data has been shown to enhance performance of tasks such as Video Representation Learning, solving Jigsaw Puzzles to learn image representations \citet{jigsaw}, etc. Similarly, ALBERT \cite{albert} shows that sentence-order prediction between two sentences is a useful pre-training objective to improve performance on various downstream NLP tasks. Inspired by such works, our work is one of the first to introduce and compare several novel `order-as-supervision' pre-training methods such as Permutation Classification, Skip-Clip, and Embedding Regression to enhance procedural understanding. These pre-training methods give a significant improvement of $1.6\%$ and $7-9\%$ compared to baselines across metrics on two downstream Entity-Tracking datasets, namely, NPN-Cooking dataset \cite{npn_paper} in the recipe domain and ProPara dataset \cite{propara-paper} in the open domain. Our methods also outperform SoTA LLMs in terms of Average Accuracy on ProPara. 

\section{Pre-training Methods}
\label{sec:pretrain}

We propose three new pre-training methods that help in learning sequential context for procedural texts: {\sl Permutation Classification}, {\sl Embedding Regression}, and {\sl Skip-Clip}. For all the methods, a set of recipes with the same number of steps are sampled. Such a recipe of $N$ steps can be represented by \(x = (x_{1}, x_{2}, . . ., x_{N})\).

\begin{figure*}[t]
\centering
\includegraphics[width=0.8\textwidth]{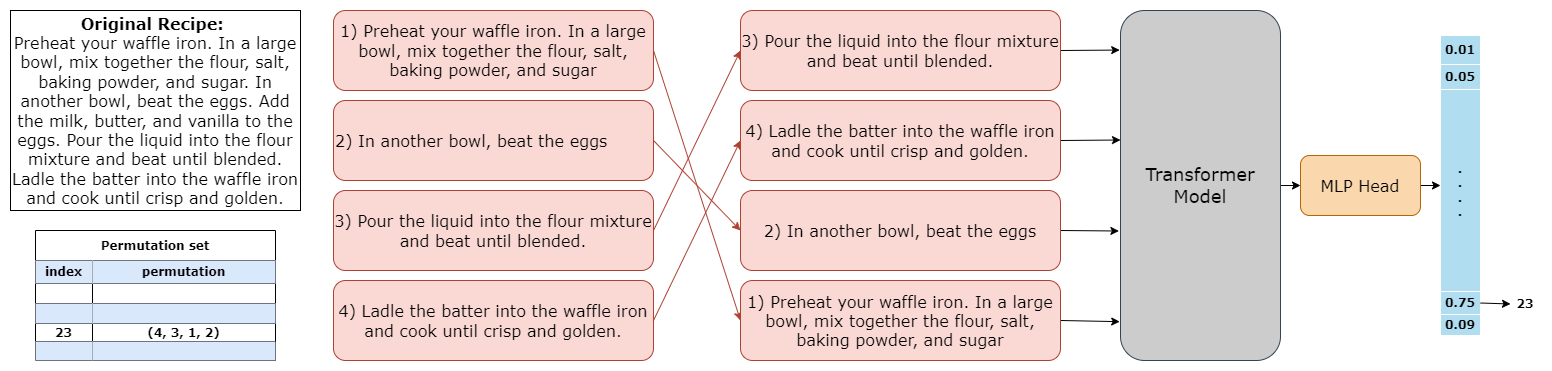}
\caption{Permutation Classification and Embedding Regression for a 4-step recipe. Recipe steps are reordered via a randomly chosen permutation from a predefined permutation set and then fed to the transformer model. The Permutation Classification Task is to predict the index of the chosen permutation which in this case is 23, and Embedding Regression Task is to predict the corresponding Lehmer/Hamming Embedding.}
\label{fig:permclassfn}
\end{figure*}

\subsection{Permutation Classification} 
In this method, the original recipe is shuffled by permuting its steps by some index permutation \(\psi_{i} = (\psi_{i1}, \psi_{i2}, . . ., \psi_{iN})\). The set of all possible permutations \(\psi^{*}\) contains \(N\)! elements. If, for example, \(N\) = 9 the total number of possible permutations equals 9! = 362,880. For practical reasons, as a pre-processing step, we reduce the set of all possible permutations by sampling a set \(\psi\) of maximally diverse permutations from \(\psi^{*}\). Following \citet{jigsaw},  we iteratively include the permutation with the maximum Hamming distance to the already chosen ones. For every recipe, we select a random permutation from this set and assign its index as a label. To solve the permutation classification task, we input the permuted sequence into a transformer and use the \(<s>\) (classification token) embedding to perform sequence multi-class classification. The number of output classes is equal to the size of the permutation set. Figure \ref{fig:permclassfn} shows the Permutation Classification Architecture.

\subsection{Embedding Regression} 
Following \citet{emb_reg}, we modify the permutation classification method defined above. Thus, instead of predicting the index, we convert the permutation into an embedding vector and perform a regression task on this embedding. We experiment with 2 different embedding constructions, considering \(\psi\) as a permutation of length $N$ - Hamming and Lehmer Embedding. Hamming Embedding ($h$) is a vector of size $N^2$ formed by concatenating one-hot vectors for each value of the permutation - $h_{N.i+j} = I\{\psi(i) = j\}$, where $I$ is the Identity Function. Lehmer Embedding ($l$) is a vector of size $N$, where the value at $i^{th}$ index is the number of indices less than $i$ with a greater permutation value - \(l_{i} = \#\{j: j<i, \psi(j)>\psi(i)\}_{1 \leq i \leq N}\). E.g. for the permutation (4,3,1,2) the Hamming Embedding is (0,0,0,1,0,0,1,0,1,0,0,0,0,1,0,0) and Lehmer Embedding is (0,1,2,2). We use Mean Squared Error (MSE) as the loss function. It can be theoretically shown that minimizing this loss on the selected embeddings is equivalent to optimizing a ranking metric like Kendall’s Tau \cite{kendalltau} or Hamming \cite{hamming} distance. Figure \ref{fig:permclassfn} shows the Embedding Regression Architecture.

\subsection{Skip-Clip} 
\begin{figure*}[t]
\centering
\includegraphics[width=0.64\textwidth]{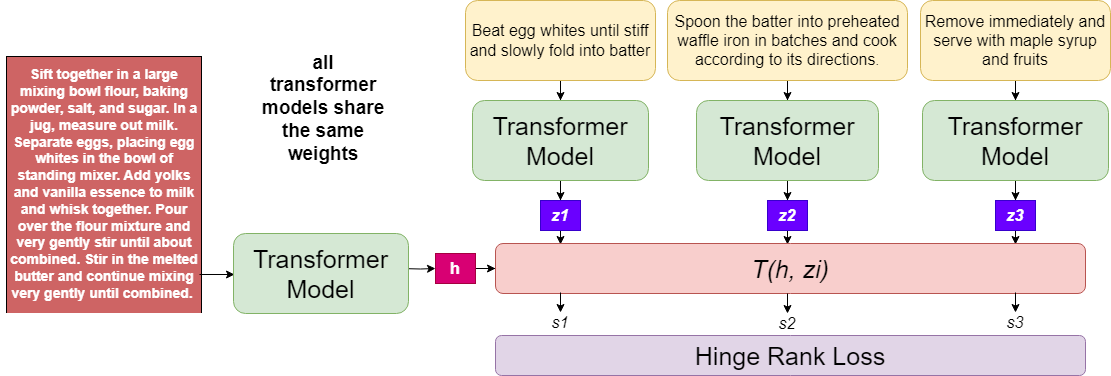}
\caption{Skip-Clip model with a 6-step context and 3 target steps. The task is to rank the target steps based on scores obtained from a scoring function and their order in the recipe using hinge rank loss.}
\label{fig:skipclip}
\end{figure*}

In this method inspired by \citet{skipclip}, given the first few steps as the context, other steps closer to the context have more similar representations to that of the context than the ones that are farther. 
Here, we sample the first \(K\) steps of a recipe with \(N\) steps, as the context \(c = (x_{1}, x_{2}, . . ., x_{K})\) and randomly sample \(M\) target steps \((x_{t_1}, x_{t_2}, . . ., x_{t_M})\), where \(t_i\) is the index in the original recipe for the $i^{th}$ target step, with \(M,K<N\) and \(t_1>K\). Using transformer model \(f\), we get latent representations of the context \(h = f(c)\) and each step in the target, \(z_{i} = f(x_{t_i})\). We also define a scoring function, \(\Gamma(h, z_i)\), e.g. cosine similarity, representing the relationship between the context and the target steps. The objective is hinge rank loss formulated as:
\(L = \sum_{i=1}^{M-1} \sum_{j=i+1}^{M} max(0, -\Gamma(h, z_i) + \Gamma(h, z_j) + \delta)\), where the constant \(\delta\) is the margin. Figure \ref{fig:skipclip} shows the Skip-Clip architecture. 

\section{Downstream tasks}
Entity Tracking consists of two sub-tasks - entity state and location tracking. Both tasks output the result for every entity at each step of the process. Entity state tracking task is a 4-way classification task that predicts if the entity is created, moved, unchanged, or destroyed at that step of the process. Entity Location tracking is formulated as a span-based question-answering problem that outputs the location of the entity at a particular step taking the entire text of the process as the input. We perform this task on  NPN-Cooking and ProPara datasets\footnote{Datasets are in the English Language}. NPN-Cooking dataset \cite{npn_paper} consists of 65,816 training, 175 development, and 700 evaluation recipes. ProPara Dataset \cite{propara-paper} consists of 488 human-authored procedures (split 80/10/10 into train/dev/test) with 81k annotations regarding changing states (existence and location) of entities in those paragraphs. The model is evaluated in 3 ways corresponding to a given entity $e$. \textbf{Category 1}: which of the three transitions - created, destroyed, or moved undergone by $e$ over the lifetime of a procedure; \textbf{Category 2}: steps at which $e$ is created, destroyed, and/or moved; and \textbf{Category 3}: the location (span of text) which indicates $e$’s creation, destruction or movement. Following \citet{TSLM}, Entity Tracking is formulated as a question-answering problem. Fine-tuning Hyperparameters are the same as in the default open-source implementation\footnote{\url{https://github.com/HLR/TSLM}}.
\section{Experiments and Results}
\subsection{Pre-training Setup}

\noindent\textbf{Parameter Initialization: }For fast convergence, we initialize the transformer in each pre-training method with RoBERTa-BASE \cite{roberta-paper}\footnote{Compute details are \textbf{\emph{in Section \ref{appendix:pretr_setup} of Appendix}}}.
\noindent\textbf{Dataset: }We use a dataset of 2.5 million+ recipes in total collected from various different sources on the internet such as Recipe1M+ dataset \cite{recipe1m+}, RecipeNLG dataset \cite{recipenlg}, datasets collected by \citet{bodhiswatta_paper} and \citet{Chandu2019StoryboardingOR}. For each recipe in the dataset, a sentence with the ingredients is also added as a step before the original recipe. The dataset is filtered to include recipes with more than 4 steps. The statistics of the dataset is shown in Table \ref{table:data_stats} \textbf{\textit{in Section \ref{appendix:pretr_setup} of Appendix}}. Permutation Classification and Embedding Regression require all recipes to have the same number of steps per recipe. Hence, we use a subset of recipes that have a certain, fixed number of steps. 
\noindent \textbf{Hyperparameters:} We pre-trained for 1 epoch using AdamW optimizer with batch size of 32, learning rate of 5e-5, weight decay of 0.01, and 500 warmup steps.

\begin{table}[!htb]
\centering
\resizebox{0.35\textwidth}{!}{%
\begin{tabular}{|l|r|r|}
\hline
\textbf{Model}                        & \multicolumn{1}{l|}{\textbf{Dev Acc}} & \multicolumn{1}{l|}{\textbf{Test Acc}} \\ \hline
NPN-Model  & -  & 51.3           \\
KG-MRC     & -  & 51.6           \\
DYNAPRO    & -  & 62.9           \\
RoBERTa-BASE                         & \underline{65.07}                                 & 64.28                                  \\ \hline
$Permutation$ $Classfn.$                 & \textbf{65.48}                                 & 64.75                                  \\ \hline

$Emb_{Hamming}$ & 65.03                                 & \underline{64.92}                                  \\ 
$Emb_{Lehmer}$              & 63.96                                 & 64.29                                  \\ \hline
$Skip$-$Clip$          & 63.87                                 & \textbf{65.33}                                  \\ \hline

\end{tabular}
}
\caption{Results on NPN-Cooking Dataset. Numbers in bold and underlined are the highest and the second-highest scores, respectively.}
\label{tab:npn_results}
\end{table}

\begin{table}[t]
\centering
\resizebox{0.45\textwidth}{!}{%
\begin{tabular}{|l|c|c|c|c|c|c|}
\hline
\textbf{Model}                       & \begin{tabular}[c]{@{}c@{}}\textbf{Location}\\\textbf{Acc.}\end{tabular} & \begin{tabular}[c]{@{}c@{}}\textbf{Status}\\\textbf{Acc.}\end{tabular} & \begin{tabular}[c]{@{}c@{}}\textbf{Cat1}\\\textbf{Acc.}\end{tabular} & \begin{tabular}[c]{@{}c@{}}\textbf{Cat2}\\\textbf{Acc.}\end{tabular} & \begin{tabular}[c]{@{}c@{}}\textbf{Cat3}\\\textbf{Acc.}\end{tabular} & \begin{tabular}[c]{@{}c@{}}\textbf{Avg.}\\\textbf{Cat.}\\\textbf{Acc.}\end{tabular} \\ \hline
Rule-based    & -  & - & 57.14         & 20.33          & 2.4     & 26.62       \\
Feature-based & -  & - & 58.64         & 20.82          & 9.66    & 29.71       \\
ProLocal      & -  & - & 62.7          & 30.5           & 10.4    & 34.53       \\
ProGlobal     & -  & - & 63            & 36.4           & 35.9    & 45.1       \\
EntNet        & -  & - & 51.6          & 18.8           & 7.8     & 26.07       \\
QRN           & -  & - & 52.4          & 15.5           & 10.9    & 26.27       \\
RoBERTa-BASE                          & 56.27                                  & 65.71                                & 71.33                              & 31.78                              & 34.05                   & 45.72           \\ \hline
$Permutation$ $Classfn.$                & 57.71                                  & \textbf{70.57}                       & \textbf{73.72}                     & \textbf{43.16}                        & 32.72                        & \textbf{49.87}      \\ \hline


$Emb_{Hamming}$ & 53.05 & 61.36 & 68.5 & 30.43 & \textbf{36.16}                   & 45.03           \\ 
$Emb_{Lehmer}$ & \textbf{60.61}                                  & \underline{68.11}                                & \underline{73.3}                               & \underline{38.82}                              & \underline{35.05}                   & \underline{49.06}           \\ \hline
$Skip$-$Clip$                  & \underline{58.19}&63.4&66.94&34.46&32.73                       & 44.71       \\ \hline
\end{tabular}
}
\caption{Results on ProPara Dataset. Numbers in bold and underlined are the highest and the second-highest scores respectively.}
\label{tab:propara_results}
\end{table}
\subsection{Fine-tuning}
We fine-tune and evaluate models pre-trained using techniques mentioned in Section \ref{sec:pretrain} on ProPara and NPN-Cooking Datasets. Note that Embedding Regression has two variants based on the type of embedding used - Hamming Embedding ($Emb_{Hamming}$) and Lehmer Embedding ($Emb_{Lehmer}$). We use hyperparameter grid search on development sets corresponding to each pre-training variant to get the best set of hyperparameters, as mentioned \textbf{\emph{in Section \ref{appendix:finetuning} of Appendix}}.



\subsection{Baselines}
\textbf{NPN-Cooking Dataset: }We use Neural Process Network Model (\textbf{NPN-Model}) \cite{npn_paper}, \textbf{KG-MRC} \cite{KGMRC}, \textbf{DYNAPRO} \cite{dynapro}, and RoBERTa-BASE \cite{roberta-paper} as baselines.

\noindent \textbf{ProPara Dataset: }We use a \textbf{rule-based} method called \textbf{ProComp} \cite{rule-based-procomp}, a \textbf{feature-based} method \cite{propara-paper} using Logistic Regression and CRF, \textbf{ProLocal} \cite{propara-paper}, \textbf{ProGlobal} \cite{propara-paper}, \textbf{EntNet} \cite{entnet}, \textbf{QRN} \cite{QRN}, and RoBERTa-BASE \cite{roberta-paper} as baselines.

\subsection{Analysis of Results}
Table \ref{tab:npn_results} shows results of proposed methods and baselines on NPN-Cooking Dataset. We see that all proposed methods perform better than baselines w.r.t Test Accuracy. Permutation Classification gives the best dev set result, but falls behind on the test set, as classification on $100$ classes leads to overfitting. Skip-Clip gives best test accuracy, with an improvement of $1.6\%$ compared to RoBERTa-BASE, suggesting that predicting next step from a given context helps in Entity Tracking in Recipe Domain. $Emb_{Hamming}$ gives the second-highest test accuracy, showing that predicting permutation as an embedding is useful for this task.

Table \ref{tab:propara_results} shows results of proposed methods and baselines on ProPara. RoBERTa-BASE is the best baseline. Most proposed methods beat baselines. Skip-Clip does not perform as well, suggesting that this pre-training method of predicting a future step in recipes does not transfer to open domain. Permutation Classification and Embedding Regression perform much better. $Emb_{Lehmer}$ performs better on 5 out of 6 metrics compared to $Emb_{Hamming}$. Permutation Classification has the best Status Accuracy and Average Category Score and gives an improvement of $7.4\%$ and $9\%$ respectively compared to RoBERTa-BASE, showing that predicting a permutation helps in a task in another domain. 

\noindent \textbf{Comparison with LLMs: }We compare with the following LLMs \textit{\textbf{in Table \ref{tab:propara-llms} in Section \ref{appendix:analysis} of Appendix}} - (1) Open-source LLMs such as Falcon-7B-Instruct (instruction-fine-tuned Falcon-7B) \cite{falcon}, Llama 2-7B-Chat (instruction-fine-tuned Llama 2-7B) \cite{llama2} (2) OpenAI's GPT-3.5 \cite{gpt35turbo}.
The LLMs are used in a 1-shot and 3-shot In-Context Learning Setting  \cite{icl}. Table \ref{tab:propara-llms} shows that - (1) even though Falcon-7B-Instruct and Llama 2-7B-Chat have almost 14x the number of parameters compared to the proposed permutation-based methods, they perform considerably worse in comparison (2) the proposed methods outperform GPT-3.5 in 1-shot setting across all metrics, and GPT-3.5 in 3-shot setting across 3 out of 4 metrics, even though the number of parameters and pre-training data is just a small fraction of that of GPT-3.5.

Additionally, we compare predictions of Permutation Classification and well-performing baseline RoBERTa-BASE on a procedure \textbf{\emph{in Table \ref{tab:propara_preds} in Section \ref{appendix:analysis} of Appendix}}. We infer that Permutation Classification is able to better predict the step when an entity ceases to exist, compared to the baseline. 

\section{Combination of different pre-training strategies}

In this section, we explore sequential combinations of pre-training strategies. As Permutation Classification performs consistently well, we experiment with one of either Skip-Clip, $Emb_{Hamming}$, or $Emb_{Lehmer}$, followed by Permutation Classification.

\begin{table}[!htb]
\centering
\resizebox{0.5\textwidth}{!}{%
\begin{tabular}{|l|r|}
\hline
\textbf{Model}                         & \multicolumn{1}{l|}{\textbf{Test Acc}} \\ \hline
$Permutation$ $Classfn.$                             & 64.75                                  \\ \hline

$Emb_{Hamming}$                      & 64.92                                  \\ 
$Emb_{Lehmer}$                                   & 64.29                                  \\ 
$Skip$-$Clip$                               & \textbf{65.33}                                  \\ \hline
$Emb_{Hamming} + Permutation$ $Classfn.$                      &  61.88       \\  
$Emb_{Lehmer} + Permutation$ $Classfn.$                                   & 62.66                    \\ 
$Skip$-$Clip + Permutation$ $Classfn.$                               &          0.01               \\ \hline

\end{tabular}
}
\caption{Results of sequential combination of different pre-training strategies on NPN-Cooking Dataset.}
\label{tab:npn_results_comb}
\end{table}

\begin{table}[!htb]
\centering
\resizebox{0.5\textwidth}{!}{%
\begin{tabular}{|l|c|c|c|c|}
\hline
\textbf{Model}                        & \begin{tabular}[c]{@{}c@{}}\textbf{Cat1}\\\textbf{Acc.}\end{tabular} & \begin{tabular}[c]{@{}c@{}}\textbf{Cat2}\\\textbf{Acc.}\end{tabular} & \begin{tabular}[c]{@{}c@{}}\textbf{Cat3}\\\textbf{Acc.}\end{tabular} & \begin{tabular}[c]{@{}c@{}}\textbf{Avg.}\\\textbf{Cat.}\\\textbf{Acc.}\end{tabular} \\ \hline

$Permutation$ $Classfn.$                               & \textbf{73.72}                     & \textbf{43.16}                        & 32.72                        & \textbf{49.87}      \\ \hline

$Emb_{Hamming}$ & 68.5 & 30.43 & \textbf{36.16}                   & 45.03           \\ 
$Emb_{Lehmer}$                                & 73.3                               & 38.82                              & 35.05                   & 49.06           \\ 
$Skip$-$Clip$ &66.94&34.46&32.73                       & 44.71       \\ \hline

$Emb_{Hamming} + Permutation$ $Classfn.$ & 63.14 & 22.74 & 34.94 & 40.27  \\ 
$Emb_{Lehmer} + Permutation$ $Classfn.$                                & 59.89 & 23.95 & 33.83 & 39.22                              \\ 
$Skip$-$Clip + Permutation$ $Classfn.$ &  44.63 & 12.44 & 5.44 & 20.84 \\ \hline

\end{tabular}
}
\caption{Results of sequential combination of different pre-training strategies on ProPara Dataset.}
\label{tab:propara_results_comb}
\end{table}

Tables \ref{tab:npn_results_comb} and \ref{tab:propara_results_comb} reveal combinations of pre-training strategies being inferior to using individual strategies, possibly because these strategies use different supervision cues. For instance, while Permutation Classification treats each permutation as an independent target class, Skip-Clip pushes representations of nearby steps closer, and vice versa. Hence, Skip-Clip + Permutation Classification performs poorly. $Emb_{Lehmer}$, unlike Permutation Classification, uses distance between each step before and after permuting as target encoding, hence, different permutations are not independent, making methods slightly inconsistent. $Emb_{Hamming}$, like Permutation Classification, has different targets for each permutation, but has a larger target vector than PC. Hence, $Emb_{Hamming}$ + Permutation Classification is reasonably good, but is inferior to each.

\section{Conclusion}
Our work is one of the first to propose order-based in-domain pre-training for procedural data to enhance Entity Tracking performance. We introduce 3 pre-training tasks - Permutation Classification, Embedding Regression, Skip-Clip to learn sequential nature of procedures. Skip-Clip performs best on the in-domain NPN-Cooking Task, while Permutation Classification and Embedding Regression perform best on the open-domain ProPara Task. We believe such methods could be extended to procedures in E-Manuals, manufacturing guides, etc.

\section*{Limitations}

\begin{enumerate}
    \item Our work focuses on recipes as a type of procedural text. It would require further study to see if the results can be generalized to other types of procedural text such as science processes or how-to guides.
    \item The Entity Tracking Task is only one aspect of understanding procedural text. Other aspects such as identifying entities and their attributes, and understanding causal relationships between entities may also be important for some applications.
    \item Our work evaluates the proposed methods on two specific datasets, which may not be representative of all possible scenarios. The performance of the methods on other datasets or real-world applications may vary.
\end{enumerate}

\section*{Ethics Statement}

The proposed methodology can be used for any type of procedural text, including user-generated procedures. However, before applying the model to such procedures, it is important to consider exposure bias patterns. Additionally, the interpretability of the model's output is limited, so users should exercise caution when using it.

\bibliography{anthology,custom}

\appendix

\section*{Appendix}


\section{Introduction}

\section{Pre-training Methods}

\section{Downstream Tasks}

\section{Experiments and Results}

\subsection{Pre-training Setup}
\label{appendix:pretr_setup}

\noindent \textbf{Compute Details: }Number of trainable parameters for each pre-training method is 488,126,475. We use Tesla V100 GPUs for our experiments. Permutation Classification and Embedding Regression Methods take about 24 GPU-Hours, while Skip-Clip takes about 8 GPU-Hours (GPU-Hours is the number of GPUs used multiplied by the training time in hours).

\noindent \textbf{Pre-training Data: }The statistics of Pre-training Data is elaborated in Table \ref{table:data_stats}.

\begin{table}[!thb]
\centering
\resizebox{0.45\textwidth}{!}{%
\begin{tabular}{|l|l|l|l|}
\hline
\textbf{Dataset}                & \textbf{No. of Recipes} & \textbf{\begin{tabular}[c]{@{}l@{}}No. of words\\ (only steps)\end{tabular}} & \textbf{\begin{tabular}[c]{@{}l@{}}No. of words\\ (only ingredients)\end{tabular}} \\ \hline
\textbf{Recipe1M+}              & 1,029,720               & 137,364,594                                                                  & 54,523,219                                                                         \\ \hline
\textbf{RecipeNLG}              & 1,643,098               & 147,281,977                                                                  & 73,655,858                                                                         \\ \hline
\textbf{Majumder et al. (2019)} & 179,217                 & 23,774,704                                                                   & 3,834,978                                                                          \\ \hline
\textbf{Chandu et al. (2019)}   & 33,720                  & 26,243,714                                                                   & -                                                                                  \\ \hline
\textbf{Total}                  & 2,885,755               & 334,664,989                                                                  & 132,014,055                                                                        \\ \hline
\end{tabular}
}
\caption{Statistics of Pre-training Data}
\label{table:data_stats}
\end{table}

\subsection{Fine-tuning}
\label{appendix:finetuning}
The set of hyperparameters used for performing Grid Search are mentioned in Tables \ref{tab:hyp1} and \ref{tab:hyp2}.

\begin{table}[!thb]
\centering
\resizebox{\columnwidth}{!}{%
\begin{tabular}{|c|c|}
\hline
\textbf{Hyperparameter}     & \textbf{Set of values} \\ \hline
Number of recipe steps      & \{4, 6, 9\}            \\ \hline
Size of the permutation set & \{2, 10, 50, 100\}     \\ \hline
\end{tabular}%
}
\caption{Set of hyperparameters used for grid search for Permutation Classification and Embedding Regression}
\label{tab:hyp1}
\end{table}

\begin{table}[!thb]
\centering
\resizebox{\columnwidth}{!}{%
\begin{tabular}{|c|c|}
\hline
\textbf{Hyperparameter}     & \textbf{Set of values} \\ \hline
Number of steps used as input context      & \{3, 4\}            \\ \hline
Number of target steps & \{3, 4\}     \\ \hline
\end{tabular}%
}
\caption{Set of hyperparameters used for grid search for Skip-Clip}
\label{tab:hyp2}
\end{table}

The best set of hyperparameters obtained are as follows - (1) \textbf{Permutation Classification: }No. of recipe steps = 6, Size of permutation set = 100 (2) \textbf{Embedding Regression: }No. of recipe steps = 6, Size of permutation set = 50 (3) \textbf{Skip-Clip: }No. of steps used as input context = 4, No. of target steps = 4.

\subsection{Baselines}

\subsection{Analysis of Results}
\label{appendix:analysis}

\noindent \textbf{Comparison with LLMs: }Table \ref{tab:propara-llms} compares performance of our proposed methods with that of LLMs in 1 and 3-shot In-Context Learning setting.

\begin{table}[!thb]
\centering
\resizebox{\columnwidth}{!}{%
\begin{tabular}{l|ccc|c}
\hline
 &
  \begin{tabular}[c]{@{}c@{}}\textbf{Cat1}\\ \textbf{Acc.}\end{tabular} &
  \begin{tabular}[c]{@{}c@{}}\textbf{Cat2}\\ \textbf{Acc.}\end{tabular} &
  \begin{tabular}[c]{@{}c@{}}\textbf{Cat3}\\ \textbf{Acc.}\end{tabular} &
  \begin{tabular}[c]{@{}c@{}}\textbf{Avg. Cat}\\ \textbf{Acc.}\end{tabular} \\ \hline
Falcon-7B-Instruct (1-shot) & 50.42 & 5.42 & 0.38 & 18.74\\
Falcon-7B-Instruct (3-shot) & 48.44 & 3.15 & 1.94 & 17.84\\
Llama 2-7B-Chat (1-shot) & 47.88 & 9.74 & 6.44 & 21.35\\
Llama 2-7B-Chat (3-shot) & 51.27 & 13.98 & 11.97 & 25.74\\
GPT-3.5 (1-shot) & 53.25 & 24.66 & 11.37 & 29.76 \\
GPT-3.5 (3-shot) & 62.43 & 34.66 & 15.81 & 37.63 \\
\hline
$Permutation$ $Classfn.$                              & \textbf{73.72}                     & \textbf{43.16}                        & 32.72                        & \textbf{49.87}      \\ \hline


$Emb_{Hamming}$  & 68.5 & 30.43 & \textbf{36.16}                   & 45.03           \\ 
$Emb_{Lehmer}$                 & \underline{73.3}                               & \underline{38.82}                              & \underline{35.05}                   & \underline{49.06}           \\ \hline
$Skip$-$Clip$                  &66.94&34.46&32.73                       & 44.71       \\ \hline
\end{tabular}%
}
\caption{Results on the ProPara Dataset - LLMs vs. proposed permutation-based methods}
\label{tab:propara-llms}
\end{table}

Table \ref{tab:propara_preds} shows annotated ground truth, predictions of Permutation Classification, and well-performing baseline RoBERTa-BASE for an entity in the procedure.
\begin{table*}[t]
\centering
\resizebox{0.7\textwidth}{!}{%
\begin{tabular}{l|c|c|c}
\hline
                             & Ground Truth & \begin{tabular}[c]{@{}c@{}}Permutation\\ Classification\end{tabular} & RoBERTa-BASE                \\ \hline
Procedure                    & flower       & flower                                                               & flower                      \\ \hline
(Before the process starts)  & -            & {\color[HTML]{009901} -}                                             & {\color[HTML]{009901} -}    \\
1. A seed is planted.        & -            & {\color[HTML]{009901} -}                                             & {\color[HTML]{009901} -}    \\
2. It becomes a seedling.    & -            & {\color[HTML]{009901} -}                                             & {\color[HTML]{009901} -}    \\
3. It grows into a tree.     & -            & {\color[HTML]{009901} -}                                             & {\color[HTML]{009901} -}    \\
4. The tree grows flowers.   & tree         & {\color[HTML]{009901} tree}                                          & {\color[HTML]{009901} tree} \\
5. The flowers become fruit. & -            & {\color[HTML]{009901} -}                                             & {\color[HTML]{FE0000} tree} \\
\begin{tabular}[c]{@{}l@{}}6. The fruits contain seeds\\ for new trees.\end{tabular} & - & {\color[HTML]{009901} -} & {\color[HTML]{FE0000} tree} \\ \hline
\end{tabular}%
}
\caption{Analysis of the ground truth and the predictions of Permutation Classification vs. a well-performing baseline on a sample from the ProPara Dataset.}
\label{tab:propara_preds}
\end{table*}

\end{document}